%% file: root.tex
\documentclass[letterpaper,10pt,journal,twoside]{ieeeconf}

\usepackage{times}
\usepackage{amsmath,amssymb,amsfonts}
\usepackage{bm}
\usepackage{mathtools}
\usepackage{booktabs,multirow,makecell,threeparttable}
\usepackage{graphicx}
\usepackage[table]{xcolor}
\usepackage{caption}
\usepackage{cuted}
\usepackage{adjustbox}
\usepackage{arydshln}
\usepackage[stretch=10,shrink=10]{microtype}
\usepackage[hidelinks]{hyperref}

\graphicspath{{FeatureSLAM_0_1e4/}}

\def\redc{\bf\cellcolor[HTML]{FF999A}}
\def\orangec{\it\cellcolor[HTML]{FFCC99}}
\def\yellowc{\cellcolor[HTML]{FFF8AD}}
\definecolor{lightgreen}{RGB}{210,245,210}
\definecolor{lightyellow}{RGB}{250,245,200}

\newcommand{\citep}[1]{\cite{#1}}

\title{FeatureSLAM: Feature-Enriched 3D Gaussian Splatting SLAM in Real Time}

\author{Christopher Thirgood$^{1a}$, Oscar Mendez$^{1a}$, Erin Chao Ling$^{1b}$, Jon Storey$^{2}$, and Simon Hadfield$^{1a}$%
	\thanks{$^{1a}$CVSSP, Computer Science and Electronic Engineering, University of Surrey, Guildford, Surrey, United Kingdom. {\tt\small \{c.thirgood, s.hadfield, o.mendez\}@surrey.ac.uk}}%
	\thanks{$^{1b}$Surrey Institute for People-Centered Artificial Intelligence, University of Surrey, Guildford, Surrey, United Kingdom. {\tt\small chao.ling@surrey.ac.uk}}%
	\thanks{$^{2}$Industrial 3D Robotics, Tonbridge, Kent, United Kingdom. {\tt\small jstorey@i3drobotics.com}}%
}

\begin{document}

\maketitle
\thispagestyle{empty}
\pagestyle{empty}

\begin{strip}
	\centering
	\vspace{-3mm}
	\includegraphics[width=1.0\textwidth]{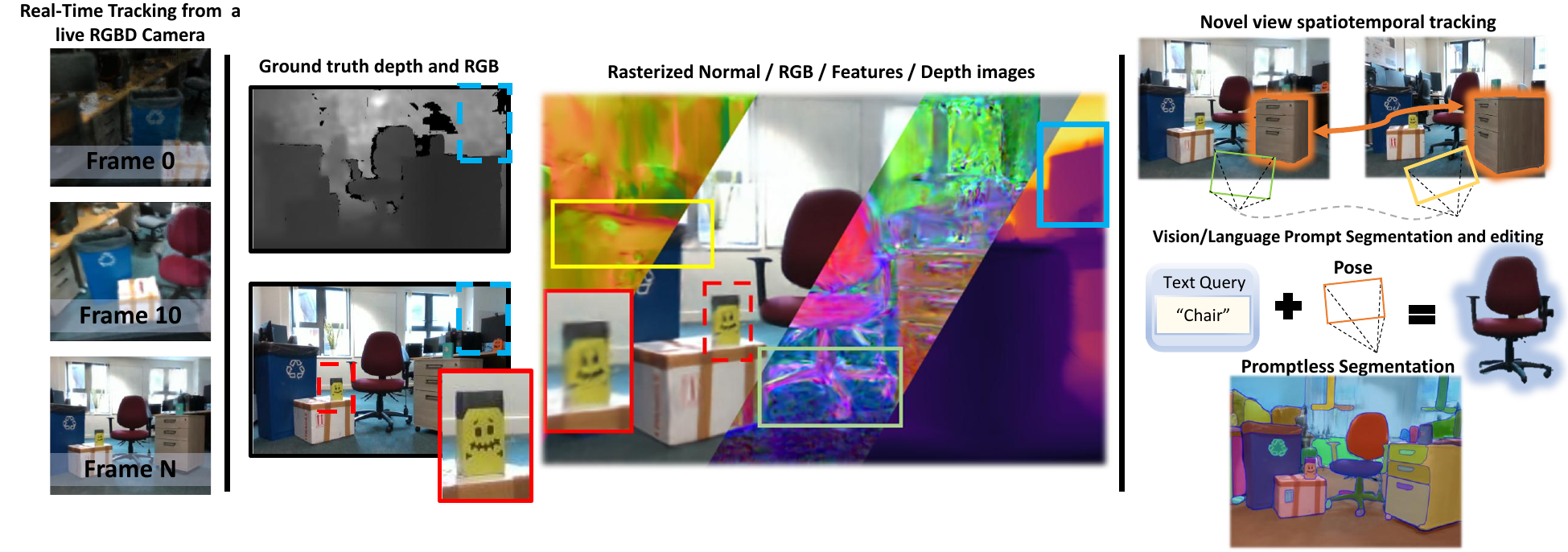}
	\vspace{-3mm}
	\captionof{figure}{Overview of FeatureSLAM. Our real-time RGB-D SLAM framework integrates foundation model feature embeddings alongside geometry, producing high-quality novel-view renderings from blurry live feeds. Left: the map is constructed online. Middle: different map modalities visualized from a novel viewpoint in our custom live viewer. Right: embedded features enable: spatio-temporal semantic tracking, promptless segmentation, and language-guided segmentation.}
	\label{fig:teaser}
\end{strip}

\input{FeatureSLAM_0_1e4/sec/0_abstract.tex}

\input{FeatureSLAM_0_1e4/sec/1_intro.tex}
\input{FeatureSLAM_0_1e4/sec/2_related_work.tex}
\input{FeatureSLAM_0_1e4/sec/3_prelim.tex}

\input{FeatureSLAM_0_1e4/sec/4_method.tex}
\input{FeatureSLAM_0_1e4/sec/5_experiments.tex}

\input{FeatureSLAM_0_1e4/sec/6_conclusion.tex}

\input{references.tex}
\end{document}

%% file: FeatureSLAM_0_1e4/sec/0_abstract.tex
\begin{abstract}
We present a real-time tracking SLAM system that unifies efficient camera tracking with photorealistic feature-enriched mapping using 3D Gaussian Splatting (3DGS).
Our main contribution is integrating dense feature rasterization into the novel-view synthesis, aligned with a visual foundation model.
This yields strong semantics, going beyond basic RGB-D input, aiding both tracking and mapping accuracy.
Unlike previous semantic SLAM approaches (which embed pre-defined class labels), FeatureSLAM enables entirely new downstream tasks via free-viewpoint open-set segmentation.
Across standard benchmarks, our method achieves real-time tracking, on par with state-of-the-art systems while improving tracking stability and map fidelity without prohibitive compute. Quantitatively, we obtain 9\% lower pose error and 8\% higher mapping accuracy compared to recent fixed-set SLAM baselines. Our results confirm that real-time feature-embedded SLAM is not only valuable for enabling new downstream applications, but also improves the performance of the underlying tracking and mapping subsystems, providing semantic and language masking results that are on par with offline 3DGS models, alongside state-of-the-art tracking, depth and RGB rendering.
\end{abstract}

%% file: FeatureSLAM_0_1e4/sec/1_intro.tex
\section{Introduction}
\label{sec:intro}
Recent SLAM systems combining novel-view synthesis with 3D Gaussian Splatting (3DGS)~\cite{3dgs} or NeRF~\cite{nerf} can produce highly detailed maps, but most depend on long offline optimization. Feature-augmented variants~\cite{feature3dgs,gs3lam,NEDS,sni} improve semantics, yet still trade real-time operation for offline training and heavy preprocessing.

Foundation models such as SAM2~\cite{Ravi2024SAM2} provide multi-scale, open-set visual features with strong generalization~\cite{vit,lseg,clip}. Coupled with grounding models~\cite{groundedSAM,groundingDINO1_5,groundingDINO}, they enable prompt-based and promptless interaction, but aligning dense hierarchical features with online, view-dependent SLAM remains challenging.

We present FeatureSLAM, a real-time 3DGS-SLAM system that embeds multi-level foundation-model features directly into the mapping and tracking loop. The result is improved tracking stability and reconstruction quality, while enabling open-set semantic interaction without offline distillation.

\textbf{In summary, our key contributions are:}
\begin{enumerate}
    \item A real-time online 3DGS-based SLAM framework that enables dense, open-set semantic mapping.
    \item A feature-guided GICP tracking approach that leverages foundation-model feature maps and camera-plane rasterization instead of traditional depth supervision.
    \item Tailored pruning, densification, and training strategies for compact feature-embedded 3DGS maps.
\end{enumerate}

%% file: FeatureSLAM_0_1e4/sec/2_related_work.tex
\section{Related Work}
\subsection{Localization in SLAM}\label{sec:lit_loc}

Early visual SLAM systems estimated camera motion and sparse landmarks using probabilistic filters, such as EKF-SLAM, or graph optimization frameworks~\cite{grisetti2010tutorial,nieto2007recursive,mallios2014scan}.  
While EKF-SLAM jointly updates both poses and landmarks, its $\mathcal{O}(n^{2})$ complexity and fragile data association limit scalability.  
Graph-based approaches decouple motion and measurement updates, improving global consistency; however, they still rely heavily on repeatable point or line features, which degrade in texture-poor environments.

Learned pose graph optimizers~\cite{learnedPGO} attempt to overcome these issues, but they still depend on robust feature tracking and often require extensive tuning.  
Geometry-based registration methods such as ICP~\cite{icp} align raw sensor data directly. Variants including point-to-plane ICP~\cite{point2planeicp}, trimmed ICP~\cite{trimmed-icp}, and Generalized ICP (GICP)~\cite{gicp} remain state-of-the-art for RGB-D and LiDAR odometry due to their speed and consistency.  
Minimalist pipelines like \textsc{KISS-ICP}~\cite{kiss-icp} and its full SLAM extension \textsc{KISS-SLAM}~\cite{kiss-slam} generalize well across sensors without per-scene tuning.  
However, all ICP-based methods require accurate geometric measurements; when depth is noisy or the field of view is narrow, tracking degrades and holes appear in reconstructed depth.

To increase robustness, several methods incorporate semantic cues.  
Techniques such as Semantic-EM ICP~\cite{semantic-emicp}, SegICP~\cite{segICP}, and SuMa++~\cite{SuMa++} augment points with class probabilities to downweight ambiguous or dynamic regions.  
SAGE-ICP~\cite{sage-icp} extends this idea to real-time LiDAR segmentation.  
These approaches highlight that semantic consistency is as critical as geometry for localization; however, they rely on preprocessed data and fixed semantic labels, which can become misaligned across frames, causing poor matching.

Inspired by this, we replace single-class labels with high-dimensional feature descriptors from open-set segmentation models such as SAM2~\cite{Ravi2024SAM2}.  
This yields viewpoint-stable features that are robust to lighting variation, occlusion, and clutter, while encoding richer semantics than traditional label-based segmentation.

\subsection{Mapping in SLAM}
\label{sec:lit_mapping}
Early dense SLAM systems such as KinectFusion~\cite{kinectfusion}, InfiniTAM~\cite{prisacariu2017infinitam}, and BundleFusion~\cite{dai2017bundlefusion} used truncated signed distance fields (TSDFs) to fuse depth into volumetric maps. These methods produced metrically accurate reconstructions but were memory-intensive and difficult to scale.
Neural implicit representations emerged as an alternative.  
NeRF~\cite{nerf} and its SLAM adaptations (iMAP~\cite{imap}, NICE-SLAM~\cite{nice-slam}, ESLAM~\cite{eslam}, Point-SLAM~\cite{pointslam}) replace TSDFs with MLP-based scene representations for photorealistic rendering.  
However, these models require per-frame backpropagation, are mostly offline, and render slowly, limiting real-world use.
Hybrid solutions such as Orbeez-SLAM~\cite{orbeez} and vMAP~\cite{kong2023vmap} decouple mapping from tracking using ORB-SLAM3~\cite{orb-slam3} for localization.  
However, loose coupling between map and tracker weakens the mutual feedback loop central to SLAM.
\vspace{-3mm}
\paragraph{3D Gaussian Splatting.}  
3DGS is a modelling technique that represents scenes as explicit collections of anisotropic Gaussians.  
GS-SLAM~\cite{yan2024gs} introduced monocular 3DGS based tracking, while SplaTAM~\cite{splatam} improved photometric alignment using silhouette constraints at the cost of slower optimization.
MonoGS~\cite{monogs} uses a gradient-based tracker on input RGB frames, and Photo-SLAM~\cite{photo-slam} incorporates ORB-SLAM3 for tracking. Yet, tracking often outpaces mapping, leading to inconsistencies, and both systems remain sensitive to hyperparameter tuning and texture-poor scenes.
RGB-D GS-ICP~\cite{gs-icp-slam} combines GICP tracking with Gaussian point positions.  
Although fast, it fails in self-similar or geometrically poor regions due to reliance on high-quality depth, and suffers from drift.
\paragraph{Feature-Distilled 3DGS.}
Recent work augments Gaussians with semantics for editing and open-vocabulary queries.  
Feature3DGS~\cite{feature3dgs} embeds SAM~\cite{sam1} or LSeg~\cite{lseg} features but requires dense training and high-dimensional feature channels.  
LangSplat~\cite{langsplat} uses language-conditioned features but with coarse boundaries due to low-dimensional embeddings.  
GaussianGrouping~\cite{gaussian_grouping} extends SAM-based masks into 3D, using DEVA~\cite{DEVA} to maintain temporal consistency.
Methods such as NEDS SLAM~\cite{NEDS}, SemGauss-SLAM~\cite{SemGauss}, and GS3 SLAM \cite{gs3lam} jointly optimize appearance, semantic features, and dense semantic label maps offline.
However, these approaches are all offline, require extensive preprocessing, fixed-resolution features and slow rendering, resulting in inconsistent semantics and poor novel view synthesis quality.

In summary, existing 3DGS SLAM methods rely solely on photometric or geometric residuals for pose estimation and optimize semantics only during mapping, or treat semantics as view-invariant scalars, ignoring directional sensitivity.  
Our approach addresses both by embedding view-conditioned SAM2 descriptors into Gaussians and jointly optimizing semantic and geometric residuals in a unified, real-time pose optimization framework.
\vspace{-2mm}

%% file: FeatureSLAM_0_1e4/sec/3_prelim.tex
\section{Preliminaries}
\label{sec:prelim}

\subsection{3DGS Rendering}
In 3D Gaussian Splatting (3DGS), each Gaussian \(\mathcal{G}_i\) is defined by its 3D position \(\bm{\mu}_i \in \mathbb{R}^3\), covariance \(\bm{\Sigma}_i \in \mathbb{R}^{3 \times 3}\), color \(\bm{c}_i \in \mathbb{R}^3\), and opacity \(\alpha_i \in [0,1]\).

A pixel color \(C\) is computed via front-to-back alpha compositing of all Gaussians, sorted by depth:
\begin{equation}
C 
= \sum_{i=1}^{N} \bm{c}_i \alpha_i \prod_{j=1}^{i-1}(1-\alpha_j)
=: \sum_{i=1}^{N} \bm{c}_i \alpha_i\, \mathcal{T}_i ,
\label{eq:blending}
\end{equation}
where \(\mathcal{T}_i = \prod_{j< i}(1-\alpha_j)\) is the accumulated transmittance.

The opacity \(\alpha_i\) for each Gaussian is modulated by its projected 2D footprint. Specifically,
\begin{equation}
\bm{\mu}_{i}^{2D} = \mathbf{K}\,\mathbf{P}\,\bm{\mu}_i,\quad
\bm{\Sigma}_{i}^{2D} = \mathbf{J}\mathbf{R}\,\bm{\Sigma}_i\,\mathbf{R}^T \mathbf{J}^T,
\label{eq:proj}
\end{equation}
where \(\mathbf{K}\) is the camera intrinsic matrix, \(\mathbf{P}\) is the world-to-camera projection matrix, \(\mathbf{R}\) is rotation, and \(\mathbf{J}\) is the Jacobian of the local affine approximation of the projection.
\vspace{-1mm}

\subsection{GICP Tracking}\label{sec:gicp}
Generalized ICP (GICP) extends classical ICP by modeling each point as a Gaussian distribution, allowing local surface geometry to influence alignment via covariance estimation. We denote source (current frame) and target (global map):
\begin{equation}
G^s = \{(\bm{\mu}^s_i, \bm{\Sigma}^s_i)\}_{i=1}^{N},\quad
G^t = \{(\bm{\mu}^t_j, \bm{\Sigma}^t_j)\}_{j=1}^{M}.
\label{eq:gicp_sets}
\end{equation}
Note that the color and opacity parameters of the 3DGS model are omitted from this process.

\paragraph{Correspondence and Gaussian error model.}

For each source Gaussian, we define the mapping $\pi$ that finds the nearest neighbor in the target set:
\begin{equation}
\pi(i) = \arg\min_j \|\bm{\mu}^s_i - \bm{\mu}^t_j\|_2.
\label{eq:correspondence}
\end{equation}
Given this correspondence, under a rigid-body transform \(T = (\mathbf{R}, \mathbf{t}) \in SE(3)\),
\begin{equation}
\bm{\mu}^s_i \mapsto \mathbf{R}\bm{\mu}^s_i + \mathbf{t},\quad
\bm{\Sigma}_i^s \mapsto \mathbf{R}\bm{\Sigma}_i^s \mathbf{R}^T.
\label{eq:rigid_transform}
\end{equation}
The residual is thus
\begin{equation}
\bm{d}_i = \bm{\mu}^t_{\pi(i)} - (\mathbf{R}\bm{\mu}^s_i + \mathbf{t}).
\label{eq:residual}
\end{equation}
Because both points are Gaussian-distributed,
\begin{equation}
\bm{d}_i \sim \mathcal{N}\bigl(\bm{0},\; \bm{\Sigma}^t_{\pi(i)} + \mathbf{R}\bm{\Sigma}^s_i\mathbf{R}^T\bigr).
\label{eq:residual_dist}
\end{equation}

\paragraph{Maximum-likelihood pose estimation.}
The optimal pose minimizes the sum of Mahalanobis distances:
\vspace{-3mm}
\begin{equation}
T^* = \arg\min_{\mathbf{R}, \mathbf{t}} \sum_{i=1}^{N}
\bm{d}_i^\top
\bigl(\bm{\Sigma}^t_{\pi(i)} + \mathbf{R}\bm{\Sigma}^s_i\mathbf{R}^T\bigr)^{-1}
\bm{d}_i.
\label{eq:gicp_pose}
\vspace{-1mm}
\end{equation}
This non-linear least-squares problem is solved iteratively using a Lie algebra parameterization of \(SE(3)\) and Gauss–Newton or Levenberg–Marquardt optimization.

%% file: FeatureSLAM_0_1e4/sec/4_method.tex
\section{Methodology}
\begin{figure*}
    \vspace{0.5em}
    \centering
    \includegraphics[width=0.8\textwidth, height=8cm]{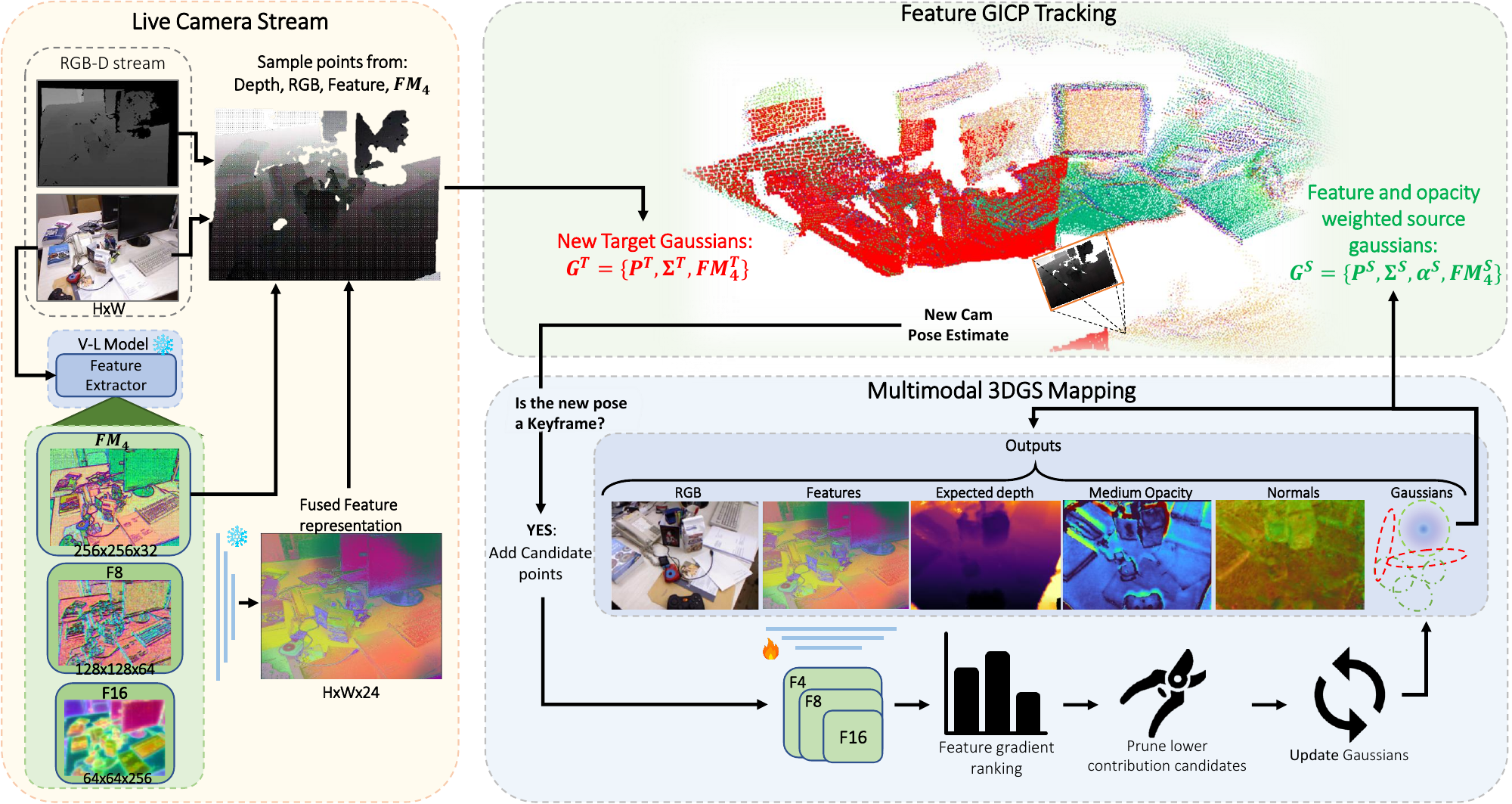}
  %

    \caption{FeatureSLAM overview, including the feature extraction, tracking and mapping modules.}    

    \label{fig2}
    \vspace{-1.0em}
\end{figure*}

Our approach enables real-time, high-resolution, feature-distilled mapping within a 3D Gaussian Splatting (3DGS) framework, facilitating semantic understanding for robotics and AR applications.
As shown in Fig.~\ref{fig2}, our pipeline consists of three main components: image feature extraction, feature-guided GICP, and feature-distilled 3DGS mapping.

\subsection{Online Feature Rasterization}


Feature distillation from foundation models into offline 3DGS has recently been shown to improve not only visual fidelity but also geometric understanding, as image encoders provide denoised, semantically rich signals that regularize view synthesis. However, all prior works rely on slower ViT/DINO-style backbones, forcing a trade-off between accuracy and throughput and precluding real-time online SLAM systems.
We instead adopt the hierarchical feature pyramid from SAM2’s Hiera-MAE encoder, which delivers multi-scale descriptors at real-time speed. Beyond efficiency, its pyramid captures complementary context (fine edges at high resolution and semantics at coarse resolution). Embedding these features in the Gaussians and decoding them jointly supplies the renderer with implicit scene priors while retaining low latency for online SLAM.

For each RGB–D frame, our encoder produces feature maps with different channel depths (256, 64 and 32) and spatial resolutions (64x64, 128x128 and 256x256, respectively), \(\{F^{(\ell)}\}_{\ell\in\{4,8,16\}}\). We bilinearly up-sample at the projected Gaussian mean \(\bm x_i\) and concatenate the 3 feature maps to form $u_i$.
A joint encoder \(\phi\) compresses \(u_i\) into a compact latent space, \(f_i=\phi(u_i)\), removing redundancies between the feature maps. Corresponding lightweight heads \(\phi^{(\ell)}\) are trained to decode back to each of the 3 feature maps. Finally, we 
augment each Gaussian \(\mathcal G_i\) in the 3DGS model with these compact feature embeddings \(f_i\in\mathbb{R}^D\).

To ensure efficient online performance, our feature autoencoder is pretrained on unrelated datasets (COCO and scannet).
To preserve this pretraining and enable efficient adaptation at runtime, we insert LoRA blocks between each head of the decoder. These LoRA blocks are frozen as zero-convolutions during pretraining. Then, during online SLAM, the main body of the autoencoder is frozen and only the LoRA blocks are unfrozen (with a gradual warmup of the learning rate). We find this approach prevents catastrophic forgetting during the online training of the autoencoder, while allowing adaptation of the learned latent-space to account for the unique artifacts induced by volumetric rendering.


\subsection{Robust Depth Rasterization}
Depth supervision is essential in 3DGS-based SLAM to suppress floating Gaussians and maintain geometric consistency. Existing systems estimate depth either as the expected value along the ray, modulated by transmittance,
\begin{equation}
    D_{\mathrm{exp}}(\mathbf{x})
    =
    \sum_{i}
    \mathcal{T}_i(\mathbf{x})\,\alpha_i(\mathbf{x})\,z_i,
    \label{eq:old_depth}
\end{equation}
or by projecting depth sensor values to each Gaussian, as in GS-ICP-SLAM~\cite{gs-icp-slam}. 
Both assume a \emph{single} depth value per Gaussian and per ray. This breaks down near surface discontinuities or slanted planes,
and introduces a strong dependency on the accuracy and completeness of the depth sensor measurements. These techniques tend to work well on synthetic data with perfect ground-truth depth, but can be unreliable for real-world operation.

Instead, we propose using structure-aware depth rasterization, where individual Gaussians induce different depths at all overlapping pixels, based on their shape and orientation. More formally, we show that under perspective projection, a 3D Gaussian produces a \emph{locally planar depth field} over its projected ellipse.

Following~\cite{Zhang2024RaDeGS}, we note that for a target pixel $\mathbf{x}$ with a corresponding ray vector $\hat{\mathbf{r}}$ we can define the maximum likelihood intersection with Gaussian $\mathcal{G}_i$ ($\boldsymbol{\mu}_i$ and $\bm{\Sigma}_i$) as
\begin{equation}
    t^\ast
    =
    \arg\max_{t}
    \exp\!\left(
    -\tfrac{1}{2}\left(
        t\hat{\mathbf{r}}-\boldsymbol{\mu}_i
    \right)^\top\!
    \bm{\Sigma}_i^{-1}
    \left(
        t\hat{\mathbf{r}}-\boldsymbol{\mu}_i
    \right)
    \right),
\end{equation}
where $t^\ast$ is the depth along the ray corresponding to the maximum density of $\mathcal{G}_i$.
This yields a closed-form solution $t^\ast = \hat{\mathbf{r}}^\top \boldsymbol{q}_i$ in ray-space, where
\begin{equation}
\boldsymbol{q}_i
=
\frac{
    \bm{\Sigma}_i^{-1}\boldsymbol{\mu}_i
}{
    \boldsymbol{\mu}_i^\top\bm{\Sigma}_i^{-1}\boldsymbol{\mu}_i
}.
\end{equation}
After perspective division and first-order expansion around $\bm{\mu}_{i}^{2D}$, the Gaussian induces a planar depth field:
\begin{equation}
    D_i(\mathbf{x})
    =
    \beta_i^\top
    \begin{bmatrix}
    \mathbf{x} - \bm{\mu}_{i}^{2D}\\
    1
    \end{bmatrix},
    \qquad
    \beta_i = [a_i,\, b_i,\, \mu_{i,z}]^\top,
\end{equation}
where $(a_i,b_i)$ are determined analytically from $\bm{\Sigma}_i$, $\boldsymbol{\mu}_i$ and the camera intrinsics. Thus, depth varies linearly over the Gaussian's 2D footprint, rather than remaining constant.
Integrating this local per-Gaussian variation into the depth rasterization of eq.~\ref{eq:old_depth}, we obtain
\begin{equation}
    D_{exp}(\mathbf{x}) 
    =
    \sum_i 
    \mathcal{T}_i(\mathbf{x})\,\alpha_i(\mathbf{x})\,D_i(\mathbf{x}).
\end{equation}
This camera-plane depth formulation eliminates the far-plane bias of GS-ICP-SLAM, preserves slanted surfaces, produces sharper silhouettes, and yields stable depth and normal gradients for optimization.

To additionally reduce bias from semi-transparent regions, we also compute the \emph{median} depth $D_{\mathrm{med}}(\mathbf{x})$ by finding the depth at which accumulated opacity first exceeds 0.5 along the ray. These are complementary: expected depth is smooth but biased; median depth preserves sharp edges.

\subsection{Depth\mbox{-}normal consistency \& regularization}
To fully exploit the camera-plane depth model, we impose losses that enforce local depth structure, surface consistency, and 
well-conditioned Gaussian geometry.

First, instead of supervising depth with only an $\ell_1$ loss, we include a patch-wise Pearson correlation loss (PCC). Given rendered depth $D$ and ground-truth depth $D^{\mathrm{obs}}$, we divide the image into $p\times p$ blocks and compute
\begin{equation}
\mathcal{L}_{\mathrm{pcc}} = 1 - \rho\bigl(D, D^{\mathrm{obs}}\bigr),
\label{eq:pcc_loss}
\end{equation}
where $\rho$ is the zero-mean, normalized cross-correlation within each block. This loss is invariant to absolute scale or bias and focuses on preserving relative depth variations (slanted planes, discontinuities, and local surface shape). This makes it especially effective in regions where Gaussian blending or transmittance leaves depth slightly biased but structurally correct.

Second, to measure surface smoothness, we derive surface normals from two complementary depth signals (expected camera plane depth and median depth) as well as from the depth sensor input ($D_{\mathrm{obs}}$). These are estimated via finite differences 
\begin{align}
    \mathbf{n}_{\mathrm{exp}}
    &= 
    \frac{
        \partial_u D_{\mathrm{exp}} 
        \times 
        \partial_v D_{\mathrm{exp}}
    }{
        \left\|
        \partial_u D_{\mathrm{exp}} 
        \times 
        \partial_v D_{\mathrm{exp}}
        \right\|},
    \\
    \mathbf{n}_{\mathrm{med}}
    &= 
    \frac{
        \partial_u D_{\mathrm{med}} 
        \times 
        \partial_v D_{\mathrm{med}}
    }{
        \left\|
        \partial_u D_{\mathrm{med}} 
        \times 
        \partial_v D_{\mathrm{med}}
        \right\|},
    \\
    \mathbf{n}_{\mathrm{obs}}
    &= 
    \frac{
        \partial_u D_{\mathrm{obs}} 
        \times 
        \partial_v D_{\mathrm{obs}}
    }{
        \left\|
        \partial_u D_{\mathrm{obs}} 
        \times 
        \partial_v D_{\mathrm{obs}}
        \right\|}.
\end{align}
%
We then introduce cosine similarity losses between each pair of normal vectors.
This encourages the depth field to form coherent planes and sharp edges, suppressing floaters and improving photometric alignment.

During online mapping, many Gaussians are initialized from noisy stereo or depth measurements and then rapidly optimized by photometric and feature losses. Without additional constraints, their covariance matrices often collapse into needle-like shapes (rank~$\approx 1$), especially along viewing rays. This is visible in the visual results shown in Fig.~\ref{exp1}. This causes unstable normals, overconfident depth, and poor alignment with the true surface tangent plane, ultimately harming tracking and pruning. This is exacerbated in real-world deployment, where missing regions of the depth map can induce sharp gradients along their borders.

To counteract this, we adopt a ranking barrier.  
Given a Gaussian with scaling vector $\mathbf{s}_i = [s_{i,x}, s_{i,y}, s_{i,z}]$, we compute normalized singular values:
\begin{equation}
p_{i,k} = \frac{s_{i,k}^2}{\sum_{j\in \{x,y,z\}} s_{i,j}^2},
\quad k \in \{x,y,z\},
\label{eq:norm_singular}
\end{equation}
and define the entropy of the scale distribution,
\begin{equation}
H_i = - \sum_{k \in \{x,y,z\}} p_{i,k}\,\log(p_{i,k}),
\label{eq:entropy}
\end{equation}
We then define $\mathrm{erank}(\Sigma_i)$ as $\exp(H_i) \in [1,3]$, where $\mathrm{erank}\approx1$ indicates a degenerate needle-shaped Gaussian, and $\mathrm{erank}\approx3$ indicates an isotropic Gaussian.

We then impose the barrier loss:
\begin{equation}
\mathcal{L}_{\mathrm{erank}}
= \lambda_e \Big[-\log\big(\mathrm{erank}(\Sigma_i) - 1 + \epsilon\big)\Big],
\label{eq:erank_loss}
\end{equation}
which encourages Gaussians to have an $\mathrm{erank}>1$, preventing collapse to 1D structures. To avoid over-regularizing Gaussians early in SLAM, $\lambda_e$ is ramped from a small initial value $\lambda_0$ to $\lambda_{\max}$ over the first $T$ iterations:
\begin{equation}
\lambda_e(t) = \lambda_0 + (\lambda_{\max} - \lambda_0)\bigl(1 - e^{-t/T}\bigr).
\label{eq:lambda_ramp}
\end{equation}

A second common mode of Gaussian collapse, which is not strongly discouraged by the above loss, is for the Gaussians to become extremely flat and disc-shaped (i.e.\ $s_{z} \ll s_{x},s_{y}$). We therefore include a thinness penalty:
\begin{equation}
\mathcal{L}_{\mathrm{thin}} =
\lambda_{\mathrm{thin}} \cdot \mathbb{E}_i\big[s_{i,3}^{\downarrow} \big],
\label{eq:thin_loss}
\end{equation}
where $s_{i,3}^{\downarrow}$ is the smallest scale after sorting $(s_x, s_y, s_z)$. This encourages a minimum Gaussian volume, preventing zero-width planar splats that are overly confident in their depth, interfering with GICP tracking or depth rendering.


Combined, the PCC depth loss, depth-normal consistency, and eRank regularization align the geometry with real surfaces, reduce drift and floater artifacts, and provide cleaner gradients for joint online mapping and tracking.

\subsection{Per-splat backward parallelization}
To sustain real-time optimization, we adopt the per-splat backward pass strategy introduced in Taming3DGS~\cite{taming3dgs}. Instead of assigning one thread per pixel (which causes heavy atomic contention during gradient accumulation), each warp iterates over splats and locally reconstructs transmittance values using precomputed checkpoints $T^*$ across subrange intervals of $P$ Gaussians:
\begin{equation}
\mathcal T^*_{kP} = T^*_{k(P-1)}\prod_{m=k(P-1)}^{kP-1} (1 - \alpha_{m}),
\label{eq:transmittance_ckpt}
\end{equation}
with the per Gaussian transmittance being recovered as
\begin{equation}
\mathcal T_{i} = \mathcal T^*_{\lfloor i/P \rfloor P} \prod_{m=\lfloor i/P \rfloor P}^{i-1} (1 - \alpha_{m}).
\label{eq:transmittance_recover}
\end{equation}
This guarantees an \emph{exact} recovery of the original transmittance,
while completely avoiding atomic operations in the backward pass.

Practically, this yields a speedup in gradient computation, enabling more inner-loop gradient steps per GICP pose update~\cite{cartgs}. However, we observe a subtle trade-off: faster photometric convergence can cause Gaussians to overfit to local color/feature residuals, drifting away from the geometry implied by the initial GICP covariance estimates. When the rendering model over-optimizes between pose updates, the accumulated error manifests as increased camera drift. As a result, while per-splat backward parallelization improves computational throughput, we cap the number of gradient steps per frame to preserve tracking stability.

\subsection{Modeling Losses}
We supervise our modeling with a range of data-driven losses on the appearance, depth, normals, and features,
\begin{equation}
\begin{aligned}
\mathcal L_{\mathrm{rgb}} &= \lambda_1\|C-C^{\mathrm{obs}}\|_1 + \lambda_2\big(1-\mathrm{SSIM}(C,C^{\mathrm{obs}})\big),\\
\mathcal L_{\mathrm{feat}} &= \lambda_f \|\hat F - F^{\mathrm{obs}}\|_1,\\
\mathcal L_{\mathrm{depth}} &= \lambda_d \|D - D^{\mathrm{obs}}\|_1 + \lambda_{\mathrm{pcc}}\,\big(1-\rho_{\mathrm{PCC}}^{(p)}(D,D^{\mathrm{obs}})\big),\\
\mathcal L_{\mathrm{nrm}} &= \lambda_n\!\left[(1\!-\!\cos\angle(\mathbf n_{\mathrm{exp}}, \mathbf n_{\mathrm{obs}})) +
(1\!-\!\cos\angle(\mathbf n_{\mathrm{med}}, \mathbf n_{\mathrm{obs}}))\right],
\end{aligned}
\label{eq:modeling_losses}
\end{equation}
where \(\rho_{\mathrm{PCC}}^{(p)}\) is a patchwise Pearson correlation (avg-pooled using blocks of size \(p\)), \(\mathbf n_{\mathrm{exp}}\) and \(\mathbf n_{\mathrm{med}}\) are rendered normals from expected and median depth, and \(\mathbf n_{\mathrm{obs}}\) is the normals from the depth sensor.
We also include the eRank barrier and ``thin'' regularizer to discourage needle-like or disc-like Gaussians while stabilizing scales.
This leads to our overall reconstruction loss,
\begin{equation}
\mathcal L=\mathcal L_{\mathrm{rgb}}+\mathcal L_{\mathrm{feat}}+\mathcal L_{\mathrm{depth}}+\mathcal L_{\mathrm{nrm}}+\mathcal L_{\mathrm{erank}}+\mathcal L_{\mathrm{thin}}.
\label{eq:total_loss}
\end{equation}

\subsection{Feature GICP tracking}
\label{sec:feat_gicp}
Classical GICP (as detailed in Sec.~\ref{sec:gicp}) aligns two point sets by minimizing a Mahalanobis distance between geometrically corresponding Gaussians. While effective on dense, noise-free depth scans, it becomes unstable in RGB–D SLAM where large texture-less regions, specular surfaces, or depth holes lead to ambiguous geometry. 

We also note a systemic discrepancy between ICP-based tracking (where every point represents a geometry measurement) and 3DGS modeling (where the geometry is aggregated across many points via volumetric rendering).
For example, it is common in 3DGS that the appearance of a single point on a complex surface might be formed through the composition of many co-located Gaussians. However, the incoming depth sensor will only produce a single point measurement at that position.

To resolve these issues, we first note that in FeatureSLAM, both our model and observations also incorporate dense, view-stable semantic information, which can help resolve misalignments. We also note that the per-Gaussian opacities can act as a proxy for the level of volumetric composition that is likely happening at a particular point. This allows us to propose a refined Tracking formulation, which is modulated to account for Gaussian visibility and semantic matching score
\begin{equation}
T^* = \arg\min_{\mathbf{R}, \mathbf{t}} \sum_{i=1}^{N}
\alpha_i(\bm{d}_i^\top
\bigl(\bm{\Sigma}^t_{\pi(i)} + \mathbf{R}\bm{\Sigma}^s_i\mathbf{R}^T\bigr)^{-1}
\bm{d}_i
+d_f),
\label{eq:feat_gicp}
\end{equation}
where $(i,\pi(i))$ is the geometric correspondence, and
\begin{equation}
d_f = \exp(-\gamma \lVert \mathbf f_i^{\,s} - \mathbf f_{\pi(i)}^{\,t} \rVert_2^2).
\label{eq:feat_dist}
\end{equation}
As in standard GICP, we linearize on $\mathfrak{se}(3)$ and solve with Gauss–Newton. 


\subsection{Keyframe Selection}
To maintain both computational efficiency and map consistency, a new keyframe is inserted only when the current camera pose deviates sufficiently from the last keyframe in either translation or rotation, or when the tracking uncertainty exceeds a predefined threshold. In particular, let $T_k$ denote the last keyframe pose and $T_c$ the current pose estimate. A keyframe is created when:
\begin{equation}
\| \mathbf t_c - \mathbf t_k \| > \tau_t
\;\; \text{or} \;\;
\angle(\mathbf R_c \mathbf R_k^\top) > \tau_r
\;\; \text{or} \;\;
\sigma_{\text{gicp}} > \tau_\sigma,
\label{eq:keyframe_crit}
\end{equation}
where $\tau_t$ and $\tau_r$ are translational and rotational motion thresholds (e.g.\ 5\,cm and $3^\circ$), and $\tau_\sigma$ is the root-mean-square GICP residual indicating pose uncertainty.  
Tracking frames contribute only to pose refinement and do not expand the 3DGS model, while keyframes are used to insert new Gaussians into the map and trigger model refinement. This balances accuracy and runtime, ensuring new geometry is fused only when it provides robust, genuinely novel viewpoints.

\subsection{Semantic gradient-based pruning \& refinement}
Foundation model features greatly enhance geometry and visual fidelity in 3DGS, but they also inflate the memory footprint and slow optimization due to their high-dimensionality.
Moreover, our Feature–GICP tracking continuously inserts new Gaussians from every keyframe's observations. Without regulation, the number of splats grows without bounds, quickly degrading rendering time and SLAM performance.

To maintain a compact but semantically meaningful scene representation, we adopt an enhanced pruning strategy via the gradient-based impact of new Gaussians on the model semantics. We refer to this process of entering new Gaussians into the 3DGS model as a "refinement round".

For each Gaussian $\mathcal{G}_i$, we define an importance score:
\begin{equation}
\psi_i = \sum_{x \in \Omega_i}
\left|
\frac{\partial \mathcal{L}_\text{rgb}}{\partial \alpha_i(x)}
\right|,
\label{eq:importance}
\end{equation}
where $\alpha_i(x)$ is the splat opacity at pixel $x$ and $\Omega_i$ is the set of pixels influenced by $\mathcal{G}_i$. This measures how the photometric error increases as the Gaussian's opacity is reduced (e.g. how important the Gaussian is for reconstruction accuracy).

We generalize this to semantic-guided pruning by also including the gradient of the feature reconstruction loss:
\begin{equation}
\psi_i =
\sum_{x \in \Omega_i}
\left(
\lambda_c \left| \frac{\partial \mathcal{L}_{\mathrm{rgb}}}{\partial \alpha_i(x)} \right|
\;+\;
\lambda_f \left|
\frac{\partial \mathcal{L}_{\mathrm{feat}}}{\partial \alpha_i(x)}
\right|
\right),
\label{eq:sem_importance}
\end{equation}
where:
$\lambda_c$ and $\lambda_f$ balance color and feature importance and $\mathcal{L}_{\mathrm{feat}} = \|\hat{F} - F^{\mathrm{obs}}\|_1$ is the feature supervision loss.
Gaussians with low $\psi_i$ contribute negligibly to appearance and semantics, making them ideal candidates for removal.

At each refinement round (triggered whenever a new tracking or mapping keyframe is added) we compute all $\psi_i$ and prune the lowest-scoring Gaussians in the $p$ percentile:
\begin{equation}
\mathcal{G}' = \left\{ \mathcal{G}_i \;\mid\; \psi_i > \mathrm{Percentile}(\psi, p)\right\}.
\label{eq:pruning_set}
\end{equation}
To preserve tracking robustness, we use a conservative removal rate during tracking keyframes ($p=10\%$) and a more aggressive threshold during mapping keyframes ($p=30\%$). This prevents excessive deletion of potential correspondences, while still keeping the map size bounded.

This strategy ensures that the retained Gaussians are those most supported by semantic gradients rather than only photometric intensity, which improves long-term geometric stability and reduces drift in low-texture regions.
At the end of a refinement round we additionally perform a standard 3DGS pruning of Gaussians that are too large or whose opacity is too low.

\subsection{Post-trajectory refinement}
While our system maps in real time, Gaussians inserted near the end of the trajectory receive fewer updates than earlier ones. After the final pose graph is fixed, we perform an optional offline refinement pass to uniformly update all Gaussians using all keyframes. 
We first recompute the 3D distance filter for the full map, ensuring consistent scale and opacity regularization across the scene, and then run a small number of additional optimization steps.
This global sweep corrects minor inconsistencies accumulated during online training and improves fine-scale geometry and feature consistency, without affecting the real-time performance of the SLAM pipeline. We report results with and without this final optimization stage. As seen in the experiments it adds around one extra minute of training to the overall time, but raises the map reconstruction accuracy greatly.

%% file: FeatureSLAM_0_1e4/sec/5_experiments.tex
\section{Experiments}
We evaluate on Replica~\cite{straub2019replica} (synthetic, clean) and TUM RGB-D~\cite{sturm2012benchmark} (real, noisy), covering both ideal and challenging operating conditions. All runs use an Intel Core i9-10900K CPU (32\,GB RAM) and an NVIDIA RTX 3090 GPU (24\,GB VRAM).

We report tracking (ATE RMSE), reconstruction (PSNR/SSIM/LPIPS), depth quality ($L_1$, precision, recall, F1), semantic quality (mIoU, mAcc), and runtime.

Baselines include: \textbf{RGB/RGB-D NVS SLAM} (Point-SLAM~\cite{pointslam}, SplaTAM~\cite{splatam}, MonoGS, Photo-SLAM, GS-ICP-SLAM~\cite{gs-icp-slam}); \textbf{closed-set semantic SLAM} (GS-SLAM~\cite{gsslam}, NIDS-SLAM~\cite{nids}, SGS-SLAM~\cite{sgs}, SNI-SLAM~\cite{sni}, NEDS-SLAM~\cite{NEDS}, SemGauss-SLAM~\cite{SemGauss}, GS3LAM~\cite{gs3lam}); and \textbf{open-set/offline feature models} (OVO-SLAM~\cite{martins2024ovo}, Feature3DGS~\cite{feature3dgs}, GS-Grouping~\cite{gaussian_grouping}, LangSplat~\cite{langsplat}, LERF~\cite{lerf2023}).

\begin{table}[!ht]
\centering
\captionsetup[table]{singlelinecheck=off}
\vspace{-2mm}
\begin{threeparttable}
\setlength{\arrayrulewidth}{1.2pt}
\resizebox{0.48\textwidth}{!}{%
\begin{tabular}{l l c c c c c c}
  \toprule
  \textbf{Category} & \textbf{Method} & \textbf{ATE$\downarrow$} & \textbf{PSNR$\uparrow$} & \textbf{SSIM$\uparrow$} & \textbf{LPIPS$\downarrow$}
  & \makecell{\textbf{L1}\\\textbf{Depth (cm)}$\downarrow$}
  & \makecell{\textbf{T. Time}\\\textbf{(mins)}$\downarrow$} \\
  \midrule

  \multirow{8}{*}{\makecell[l]{Visual\\SLAM}}
  & Point-SLAM~\cite{pointslam}      & 0.46 & 29.43 & 0.935 & 0.235  & \redc 0.44 & 144.92 \\
  & SplaTAM~\citep{splatam}          & 0.35 & 33.91 & 0.969 & 0.097 & 0.72 & 238.09 \\
  & GS-SLAM~\citep{gsslam}           & 0.50 & 34.27 & 0.975 & 0.082 & 1.16 & \phantom{00}4.01 \\
  & LoopSplat~\citep{loopsplat}      & \yellowc0.26 & 36.33 & 0.981 & 0.112 & 0.62 & 163.00 \\
  & MonoGS~\cite{monogs}             & 0.58 & \yellowc36.70 & 0.960 & 0.064 & 1.10 & 200.75 \\
  & Photo-SLAM~\cite{photo-slam}     & 0.44 & 34.13 & 0.970 & 0.099 & 2.50 & \phantom{00}1.45 \\
  & CartGS~\cite{cartgs}             & 0.48 & 34.64 & 0.970 & 0.101 & 2.30 & \phantom{00}1.45 \\
  & GS-ICP-SLAM~\citep{gs-icp-slam}  & \orangec0.16 & \orangec38.86 & \yellowc0.970 & \orangec0.049 & 4.54 & \phantom{00}1.11 \\
  \midrule

  \multirow{3}{*}{\makecell[l]{Semantic\\SLAM}}
  & SNI-SLAM~\citep{sni}             & 0.46 & 29.43 & 0.935 & 0.235 & 0.87 & 54.64 \\
  & SemGauss-SLAM~\citep{SemGauss}   & 0.38 & 34.83 & 0.978 & 0.069 & \orangec 0.53 & 140.90 \\
  & GS3LAM~\citep{gs3lam}            & 0.37 & 36.26 & \orangec0.989 & \yellowc0.052 & \yellowc{0.62} & 152.40 \\
  \midrule

  \multirow{2}{*}{\makecell[l]{Feature\\SLAM}}
  & \textbf{Ours}        & \redc\textbf{0.15} & \redc\textbf{41.22} & \redc\textbf{0.986} & \redc\textbf{0.046} & \textbf{0.73} & \textbf{\phantom{00}4.44} \\
  \cdashline{2-8}[0.4pt/1.5pt]
  & \textbf{Ours+refine} & \textbf{0.15} & \textbf{42.52} & \textbf{0.989} & \textbf{0.040} & \textbf{0.42} & \textbf{\phantom{00}5.14} \\
  \bottomrule
\end{tabular}%
}
\end{threeparttable}
\vspace{-2mm}
\caption{Camera tracking and reconstruction on Replica~\citep{straub2019replica} (average over 8 scenes).}
\label{tab:ate_replica}
\vspace{-4mm}
\end{table}

\begin{table}[!ht]
\centering
\captionsetup[table]{singlelinecheck=off}
\begin{threeparttable}
\setlength{\arrayrulewidth}{1.2pt}
\resizebox{0.48\textwidth}{!}{%
\begin{adjustbox}{trim=0pt 0pt 0pt 0pt, clip} 
\begin{tabular}{@{}lccccc@{}}
  \toprule
  \textbf{Method} & \textbf{ATE$\downarrow$} & \textbf{PSNR$\uparrow$} & \textbf{SSIM$\uparrow$} & \textbf{LPIPS$\downarrow$}
  & \makecell{\textbf{T. Time}\\\textbf{(mins)}$\downarrow$} \\
  \midrule
    Photo\mbox{-}SLAM~\citep{photo-slam} & 0.26 & 18.39      & 0.647         & 0.221         & \phantom{00}4.4 \\
    CartGS~\citep{cartgs}                & 0.27    & 18.55      & 0.677         & 0.216 & \phantom{00}4.2 \\
  \midrule
    SplaTAM~\citep{splatam}              & 5.50           & \redc{22.77} & \redc{0.867} & \redc{0.173}  & 210.2 \\
    MonoGS~\citep{monogs}                & \redc{1.48} & \yellowc{18.60}       & \orangec{0.751}          & \yellowc{0.199}         & \yellowc{130.2}  \\
    GS\mbox{-}ICP\mbox{-}SLAM~\citep{gs-icp-slam} & \yellowc{2.64} & 18.32 & 0.634 & 0.246 & \redc{\phantom{00}3.6}  \\
    \textbf{Ours}                        & \orangec{\textbf{2.05}}           & \orangec{\textbf{18.82}} & \yellowc{\textbf{0.653}} & \orangec{\textbf{0.190}} & \orangec{\textbf{\phantom{0}24.9}}  \\
    \midrule
    \textbf{Ours+refine}                        & \textbf{2.05}           & \textbf{19.22} & \textbf{0.683} & \textbf{0.187} & \textbf{\phantom{0}26.9}  \\
  \bottomrule
\end{tabular}%
\end{adjustbox}
}
\end{threeparttable}
\vspace{-2mm}
\caption{Camera tracking and reconstruction on TUM~\citep{sturm2012benchmark} (average over 3 scenes).}
\label{tab:ate_tum}
\vspace{-4mm}
\end{table}
\vspace{-2mm}

\subsection{Tracking and Reconstruction Accuracy}
Tables~\ref{tab:ate_replica} and~\ref{tab:ate_tum} present results on the synthetic Replica ~\cite{straub2019replica} and TUM-RGBD\cite{sturm2012benchmark} datasets respectively. For Replica, our method achieves SOTA tracking accuracy across all evaluated scenes (see supplementary for the per-scene breakdown), consistently outperforming coupled Gaussian SLAM systems such as the baseline method GS-ICP SLAM~\cite{gs-icp-slam} and the decoupled methods Photo-SLAM~\cite{photo-slam}/CartGS\cite{cartgs}. This improvement stems from the combination of multi-layered semantically informed feature correspondences, camera-plane depth and normal estimation, and a compact but expressive Gaussian representation. Unlike purely geometric methods, our tracking is guided not only by 3D structure but also by the high-level semantic features embedded in our model, allowing correspondences to remain stable in low-texture or repetitive regions. 

These gains are achieved while running in real time at approximately 5\,FPS, without relying on precomputed semantic masks, global bundle adjustment, or loop-closure optimization. Competing approaches that match or surpass our ATE, such as ORB-SLAM3 or loop-closure-based systems like LoopSplat~\cite{loopsplat}, generally require pose graph optimization over the entire sequence and therefore compromise real-time performance.

\begin{figure}
    \centering
    \includegraphics[width=0.455\textwidth]{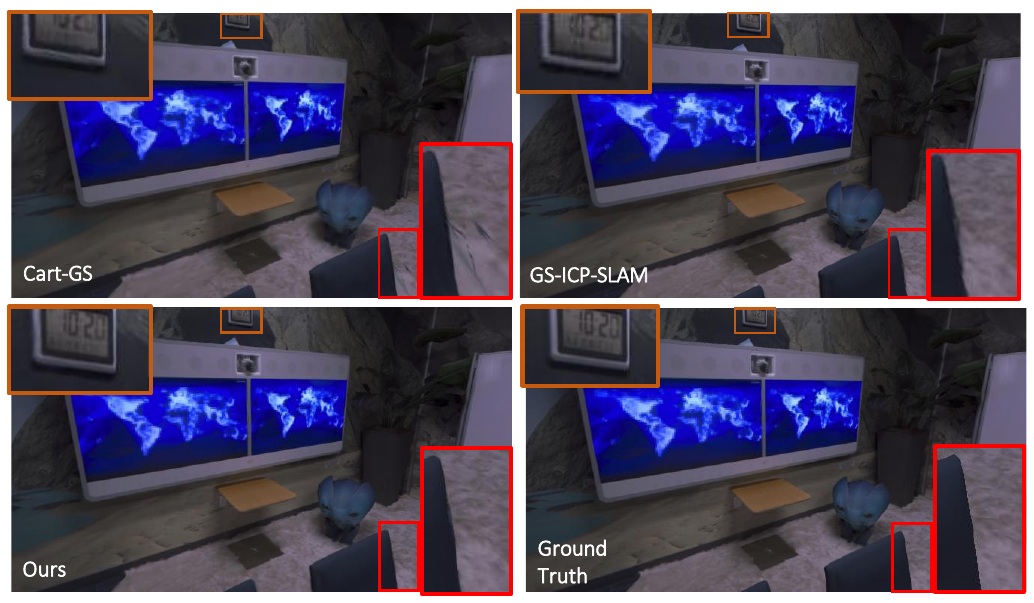}
    
    \caption{Qualitative comparison of novel-views}
    \label{exp1}
    \vspace{-1.0em}
\end{figure}

On the real-world TUM RGB-D dataset~\cite{sturm2012benchmark}, our method also achieves comparably low trajectory drift to existing coupled Gaussian-map SLAM systems. Decoupled methods like Photo-SLAM/CartGS achieve a lower ATE but provide lower quality mapping with no semantic understanding. Learned PGO methods succeed in this task but these methods are more than 7 times slower than our approach without providing improved visual results or any semantic segmentation capabilities. The baseline, GS-ICP-SLAM, when evaluated without a depth mask yields very noisy gaussians with poorly localized edges. Although fast, this method suffers in datasets where the quality of depth is poor. This is evident from the significant performance drop between Replica and TUM. In contrast, our method
offers a more robust scene representation with excellent reconstruction quality, while still operating online.
Comparative novel view reconstructions are shown in Figure~\ref{exp1} highlighting our systems higher level of detail and the absence of floater artifacts.


\subsection{Novel-View Open-Set Segmentation}
To assess open-set generalization as a downstream task, we integrate our 3D feature fields with open-vocabulary queries via GroundingDINO (see supplementary materials) as a seperate downstream task.
We compare against offline distillation-based 3DGS models: Feature3DGS~\cite{feature3dgs}, LERF~\cite{lerf2023}, and LangSplat~\cite{langsplat}. These methods require full-scene training and pre-extracted 2D features and do not run in real time. For fairness, we provide our SLAM keyframes and ground-truth poses, and train with the authors’ default settings for an equal number of iterations.
We omit prior semantic SLAM methods that rely on ground-truth label inputs and pre-determined class labels, as these cannot be compared like-for-like on the open-set segmentation task.

Table~\ref{tab:3d_segmentation_results} shows that \textbf{FeatureSLAM} matches or exceeds all baselines in zero-shot, novel-view segmentation while also running fully online. The only offline technique that performs on par with FeatureSLAM is LERF, which does well in pixel wise segmentation accuracy (+5\%). However, the MIoU score of LERF is significantly lower (-15\%).
Importantly, our training is ${\sim}10\times$ shorter than the next fastest method, and unlike some baselines, requires no dataset preprocessing for masks or CLIP embeddings.

\begin{figure}[t]
    \centering
    \includegraphics[width=\linewidth]{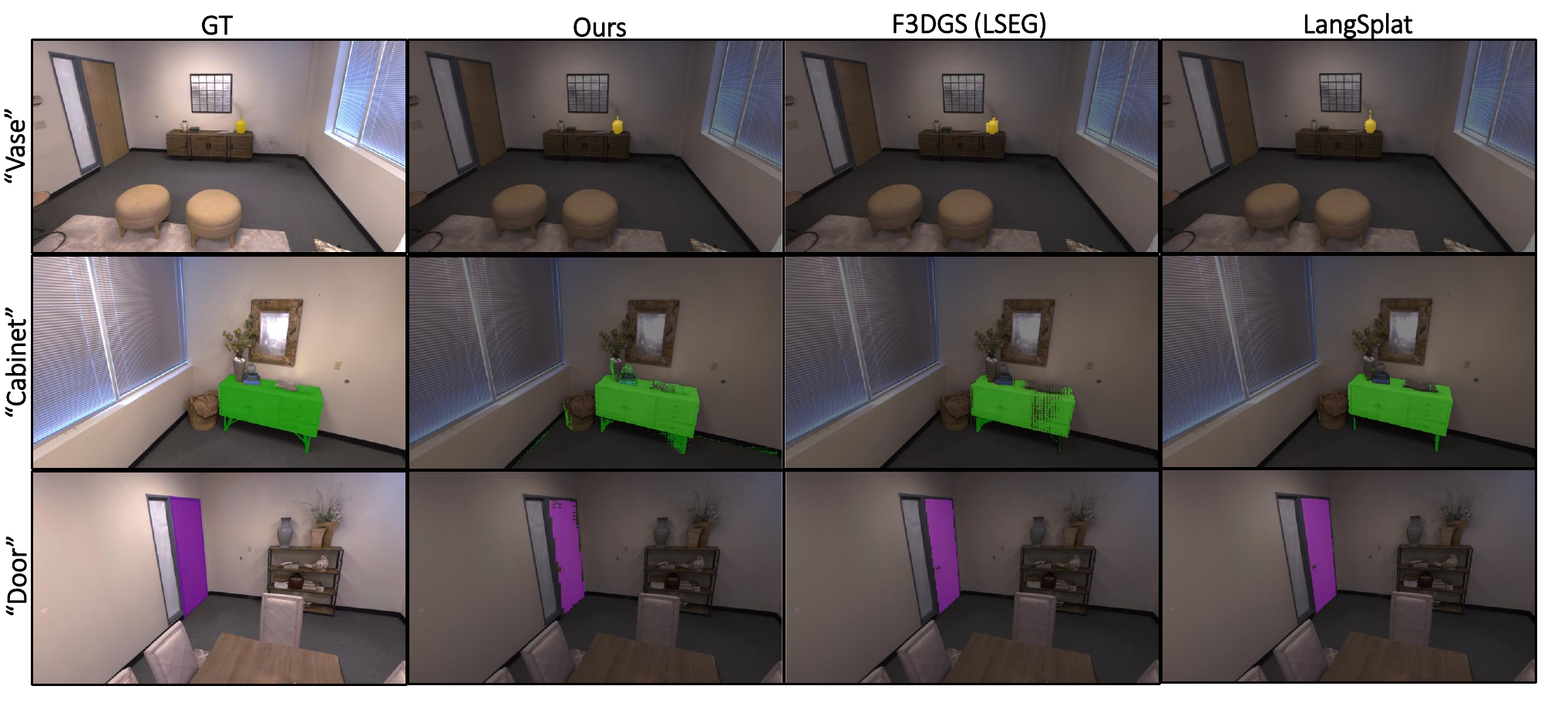}
    \caption{Examples of language prompting used with featureSLAM against Feature3DGS using LSeg and LangSplat.}
    \label{fig:lang-prompt-segs}
    \vspace{-4mm}
\end{figure}

\begin{figure}[h]
    \centering
    \includegraphics[width=\columnwidth]{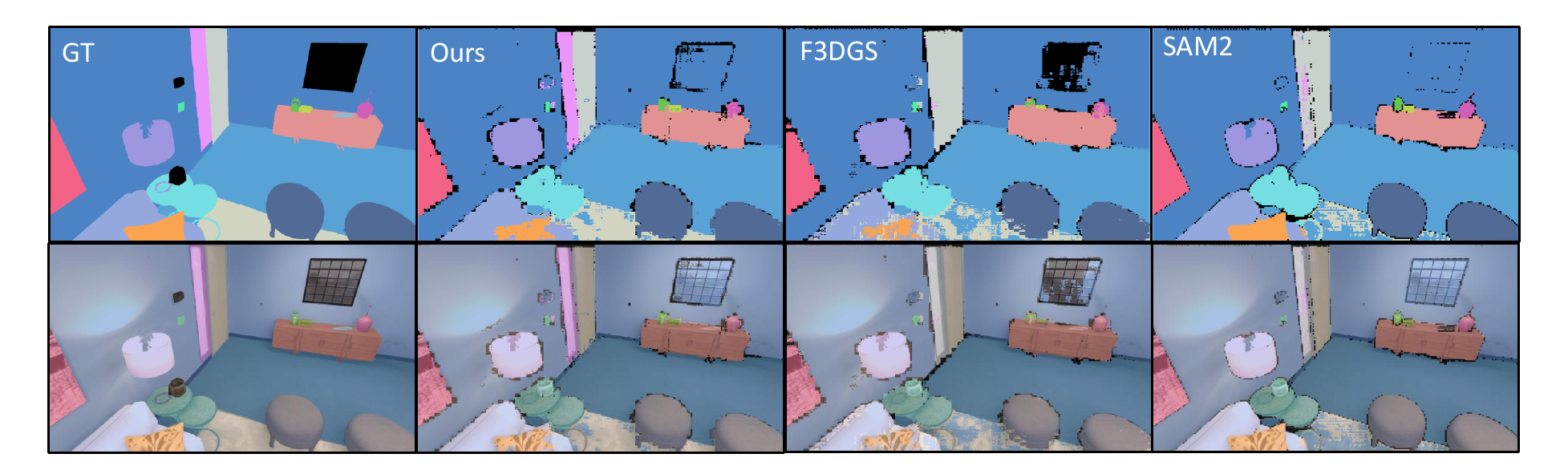}
    \caption{Promptless segmentation result with overlay visualisation from GT rgb image.}
    \label{fig:placeholder0}
\vspace{-4mm}
\end{figure}

\begin{table}[t]
\centering
\scriptsize
\begin{tabular}{@{}p{2.3cm}p{1.55cm}p{0.6cm}p{0.6cm}p{1.0cm}@{}}
\hline
\textbf{Method}          & \textbf{Backbone} & \textbf{mIoU, \%} & \textbf{mAcc, \%} & \textbf{Training Time} \\ \hline
LERF~\cite{lerf2023}               & NeRF+CLIP   & \yellowc31.2          & \redc60.7          & \phantom{0}45 min \\
LangSplat~\cite{langsplat}        & 3DGS+CLIP   & 24.7          & 42.0          & 180 min \\
Feature3DGS~\cite{feature3dgs}       & 3DGS+LSeg   & \orangec39.9          & \yellowc52.1          & 150 min  \\
\midrule
OVO-G.-SLAM~\cite{martins2024ovo}  & 3DGS+CLIP   & 22.3          & 41.1          & 138 min  \\

\textbf{FeatureSLAM} & 3DGS+GSAM2 & \redc46.3 & \orangec\textbf{55.3} & \textbf{5.4 min} \\ \hline
\end{tabular}

\caption{Comparison of downstream segmentation results on Replica dataset including offline and online feature only methods.}
\label{tab:3d_segmentation_results}
\end{table}

%% file: FeatureSLAM_0_1e4/sec/6_conclusion.tex
\vspace{-1mm}
\section{Conclusion}
This paper presented FeatureSLAM, a coupled 3DGS SLAM method capable of operating online in real time, with no preprocessing required. Our novel tracking and mapping methodology provides competitive scene fidelity while enabling open-set segmentation capabilities rivaling offline methods that use extensive preprocessing and ground-truth poses. Our real-world tests show promise for downstream tasks that current SLAM systems do not support. Future work will focus on improving robustness in dynamic scenes, improving long-horizon map consistency, and scaling the system to larger environments while preserving real-time performance.